\documentclass{article} 
\usepackage[preprint]{colm2026_conference}

\usepackage{microtype}
\usepackage{hyperref}
\usepackage{url}
\usepackage{booktabs}
\usepackage{amsmath}
\usepackage{amssymb}
\usepackage{amsthm}
\usepackage{algorithm}
\usepackage{algorithmic}
\usepackage{tikz}
\usetikzlibrary{backgrounds}
\usepackage{xcolor}

\usepackage{lineno}

\definecolor{darkblue}{rgb}{0, 0, 0.5}
\hypersetup{colorlinks=true, citecolor=darkblue, linkcolor=darkblue, urlcolor=darkblue}

\title{Check Token: Real-Time Self-Verification and Precise Truncation in LLM Reasoning}

\author{
Dingding\thanks{Corresponding author: \texttt{dignfei@gmail.com}} \& Yongkang Zhang \\
Alibaba Group \\
\And
Yuhao Liao, Zijian Zeng \& Huiming Yang \\
Tsinghua University
}

\newtheorem{theorem}{Theorem}
\newtheorem{proposition}{Proposition}

\newtheorem{corollary}{Corollary}
\newtheorem{definition}{Definition}

\newcommand{\CHECK}{\texttt{<CHECK>}}

\begin{document}

\ifcolmsubmission
\linenumbers
\fi

\maketitle

\begin{abstract}
ECC memory embeds 8 parity bits for every 64 data bits and automatically detects and corrects errors on each read. The parity bits carry no data and only safeguard integrity, at $\sim$12.5\% overhead. Yet the reasoning chains of large language models lack such built-in self-verification: once an error occurs it propagates along the chain, and existing methods can only verify externally after generation completes. We propose the \textbf{check token}, establishing built-in self-verification for language model generation streams for the first time: a functional marker \CHECK{} is added to the vocabulary, and the model triggers self-checking (analysis, localization, truncation, rewriting) by outputting it at any position. The check token carries no reasoning content (discarded after triggering) and only safeguards reasoning correctness, directly corresponding to the role of ECC parity bits, also at only $\sim$13\% overhead. \textbf{Speculative forking} further eliminates false-positive latency, and \textbf{segmented reward} makes trigger timing end-to-end learnable. Experiments (Qwen3-32B / Qwen3-Next) show that the check token achieves $+10.8$\,pp improvement on HMMT25 at $1.13\times$ overhead, with token efficiency $88\times$ that of Best-of-$8$, and the precise truncation advantage monotonically increases with chain length.
\end{abstract}

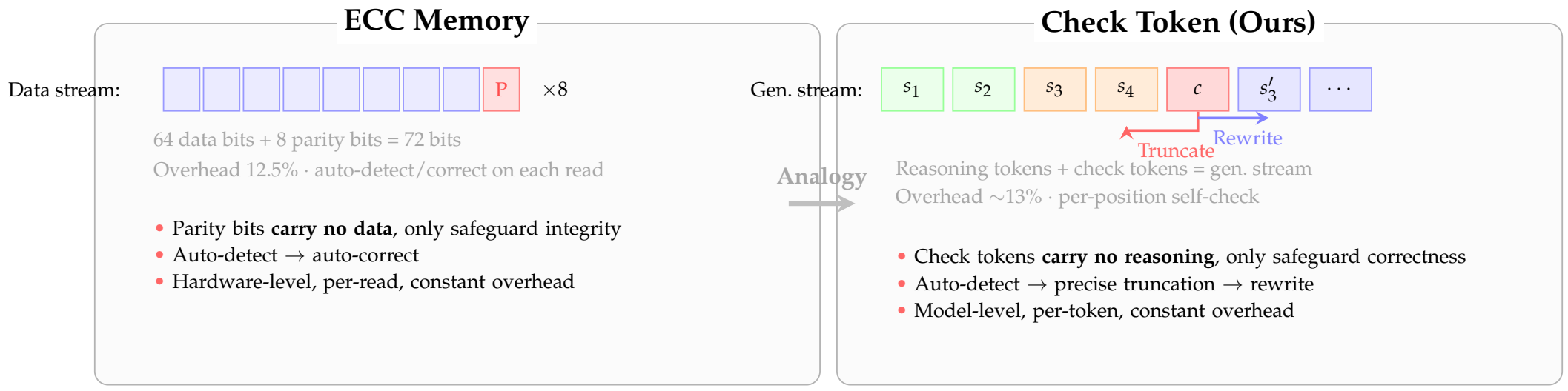
\begin{figure*}[h]
\centering
\resizebox{\textwidth}{!}{%
\begin{tikzpicture}[
    node distance=0.15cm,
    bit/.style={rectangle, draw, minimum width=0.38cm, minimum height=0.45cm, font=\tiny, inner sep=0pt},
    dbit/.style={bit, fill=blue!8, draw=blue!40},
    pbit/.style={bit, fill=red!12, draw=red!50},
    tok/.style={rectangle, draw, minimum width=0.65cm, minimum height=0.45cm, font=\tiny, inner sep=1pt},
    oktok/.style={tok, fill=green!8, draw=green!40},
    errtok/.style={tok, fill=orange!15, draw=orange!50},
    chktok/.style={tok, fill=red!15, draw=red!50},
    newtok/.style={tok, fill=blue!10, draw=blue!40},
    bigbox/.style={rectangle, rounded corners=4pt, draw=gray!60, fill=gray!3, minimum height=3.8cm, text width=7.2cm, inner sep=6pt},
    titlebox/.style={font=\small\bfseries, fill=white, inner sep=2pt},
    lbl/.style={font=\tiny},
    annot/.style={font=\tiny, text=gray!70},
    arr/.style={->, thick, >=stealth},
    darr/.style={->, thick, >=stealth, dashed, gray!60}
]

\node[bigbox, anchor=north west] (eccbox) at (0, 0) {};
\node[titlebox] at (3.6, 0) {ECC Memory};

\node[lbl, anchor=east] at (0.4, -0.7) {Data stream:};
\foreach \i in {1,...,8} {
    \node[dbit] (d\i) at ({0.5+\i*0.42}, -0.7) {};
}
\node[pbit] (p1) at ({0.5+9*0.42}, -0.7) {\textcolor{red!70}{P}};
\node[lbl, anchor=west] at ({0.5+9.7*0.42}, -0.7) {$\times 8$};

\node[annot, anchor=west] at (0.5, -1.25) {64 data bits + 8 parity bits = 72 bits};
\node[annot, anchor=west] at (0.5, -1.55) {Overhead 12.5\% $\cdot$ auto-detect/correct on each read};

\node[lbl, anchor=north west, text width=6.5cm] at (0.5, -1.95) {
\textcolor{red!60}{\textbullet} Parity bits \textbf{carry no data}, only safeguard integrity\\[1pt]
\textcolor{red!60}{\textbullet} Auto-detect $\to$ auto-correct\\[1pt]
\textcolor{red!60}{\textbullet} Hardware-level, per-read, constant overhead
};

\node[bigbox, anchor=north west] (ctbox) at (7.8, 0) {};
\node[titlebox] at (11.4, 0) {Check Token (Ours)};

\node[lbl, anchor=east] at (8.2, -0.7) {Gen.\ stream:};
\node[oktok] (t1) at (8.6, -0.7) {$s_1$};
\node[oktok] (t2) at (9.35, -0.7) {$s_2$};
\node[errtok] (t3) at (10.1, -0.7) {$s_3$};
\node[errtok] (t4) at (10.85, -0.7) {$s_4$};
\node[chktok] (tc) at (11.6, -0.7) {$c$};
\node[newtok] (t3p) at (12.35, -0.7) {$s_3'$};
\node[newtok] (t4p) at (13.1, -0.7) {$\cdots$};

\draw[arr, red!60] (tc.south) -- ++(0, -0.2) -- ++(-0.75, -0.0) node[lbl, below, pos=0.3] {Truncate} -- ++(0, 0.05);
\draw[arr, blue!50] (11.6, -1.0) -- ++(0.75, 0) node[lbl, below, pos=0.7] {Rewrite};

\node[annot, anchor=west] at (8.3, -1.55) {Reasoning tokens + check tokens = gen.\ stream};
\node[annot, anchor=west] at (8.3, -1.85) {Overhead $\sim$13\% $\cdot$ per-position self-check};

\node[lbl, anchor=north west, text width=6.5cm] at (8.3, -2.25) {
\textcolor{red!60}{\textbullet} Check tokens \textbf{carry no reasoning}, only safeguard correctness\\[1pt]
\textcolor{red!60}{\textbullet} Auto-detect $\to$ precise truncation $\to$ rewrite\\[1pt]
\textcolor{red!60}{\textbullet} Model-level, per-token, constant overhead
};

\draw[arr, line width=1.5pt, gray!50] (7.3, -1.9) -- (8.0, -1.9) node[midway, above, font=\scriptsize\bfseries, text=gray!70] {Analogy};

\end{tikzpicture}%
}
\caption{Analogy between ECC memory and check tokens. ECC embeds 8 parity bits per 64 data bits (12.5\% overhead), auto-detecting and correcting on each read. Check tokens embed self-verification into LLM generation streams ($\sim$13\% overhead), triggering self-checking, precise truncation, and rewriting at any position. Both share the same design philosophy: the verification mechanism carries no data/reasoning content, only safeguards integrity/correctness, with constant overhead independent of stream length.}
\label{fig:teaser}
\end{figure*}

\section{Introduction}
\label{sec:intro}

ECC memory embeds 8 parity bits for every 64 data bits and automatically detects and corrects errors on each read. The parity bits carry no data, only safeguard integrity, at merely $\sim$12.5\% overhead. This idea of ``building self-verification into the data stream'' is the cornerstone of hardware reliability. Yet the reasoning chains of large language models \citep{NEURIPS2022_9d560961} still lack an analogous mechanism. Reasoning chains have grown from a few steps to hundreds of steps, and an error at step $j$ invalidates all subsequent $N-j$ steps (Theorem~\ref{thm:info_retention}). Existing error-correction paradigms rely on post-generation external verifiers \citep{lightman2024lets} or majority voting \citep{wang2023selfconsistency}, equivalent to ``running the entire program and then checking from scratch,'' unable to cut losses during generation. Recent o1/DeepSeek-R1 models have exhibited emergent natural-language self-correction, but trigger words are constrained by grammar to appear only at sentence boundaries, with effective frequency of only $1/\bar{L}_{\text{sent}}$, far from the density of ``per-token verification.'' To our knowledge, no existing method simultaneously possesses all five capabilities: precise localization, error analysis, truncation, learned triggering, and general reasoning verification (Table~\ref{tab:comparison}).

We propose the \textbf{check token mechanism}, establishing built-in self-verification for language model generation streams for the first time. Just as ECC serves memory, the check token aims to become the reliability infrastructure for AI reasoning: independent of external verifiers, requiring no architectural changes, with controllable overhead ($\sim$13\%), embeddable in any autoregressive LLM. Our contributions are:
\begin{enumerate}
    \item \textbf{Built-in self-verification infrastructure for LLM reasoning}. We add \CHECK{} to the vocabulary; the model triggers self-checking at any position (analysis $\to$ localization $\to$ truncation $\to$ rewriting), with four-level fallback localization (only $3.3\%$ complete failure). The mechanism is \textbf{self-contained} (no external verifier), \textbf{zero-interference} (discarded after triggering, no impact on original reasoning), and \textbf{architecture-agnostic} (vocabulary extension $+$ segmented reward applies to any autoregressive LLM). Just as ECC parity bits carry no data and only safeguard integrity, check tokens carry no reasoning and only safeguard correctness. AIME25/HMMT25/LiveCodeBench achieve $+9.2$/$+10.8$/$+7.2$\,pp improvements, with token efficiency $88\times$ that of Best-of-$8$.
    \item \textbf{Breaking the structural bottleneck of natural-language self-correction}. The trigger words of o1/DeepSeek-R1 are grammatically constrained to appear only at sentence boundaries; check tokens are unconstrained, increasing effective trigger frequency by $\sim\bar{L}_{\text{sent}}\times$, reducing the error length required to achieve the same detection probability by $\bar{L}_{\text{sent}}\times$ (Theorem~\ref{thm:correction_prob}).
    \item \textbf{Speculative forking inference}. False-positive latency is zero; effective latency drops from serial $\sim$11\% to $\sim$3.5\%, with both modes producing identical outputs.
    \item \textbf{Complete theoretical guarantees}. Three theorems: detection probability approaches $1$ exponentially; precise truncation preserves $\Theta(N)$ correct reasoning; joint bound reveals that the marginal value of improving $a$ exceeds that of improving $p$.
    \item \textbf{Segmented reward mechanism}. The optimal trigger threshold $q^*$ has a closed-form solution, and relative token overhead is independent of $N$ (theoretical $\sim11\%$, measured $\sim13\%$).
\end{enumerate}

\section{Related Work}
\label{sec:related}

\paragraph{LLM Reasoning and Self-Correction}
CoT \citep{NEURIPS2022_9d560961} enables step-by-step reasoning, but once an error occurs subsequent steps fail accordingly. Tree-of-Thoughts \citep{NEURIPS2023_271db992} mitigates this through multi-path search but incurs $O(b^d)$ overhead; Self-Refine \citep{NEURIPS2023_91edff07} iteratively improves through multiple rounds but has limited effectiveness on reasoning tasks \citep{ICLR2024_8b4add8b}; Self-Verification \citep{weng-etal-2023-large} operates only after generation. All these methods belong to the ``post-generation correction'' paradigm, losing $\Theta(N)$ already-generated information as chain length $N$ increases (Theorem~\ref{thm:info_retention}).

\paragraph{Process Rewards and Verification}
PRM \citep{lightman2024lets} and Math-Shepherd \citep{wang-etal-2024-math} require separately trained external verifiers and only provide post-hoc scoring. Our segmented reward internalizes quality signals into the generation model: rewards are jointly determined by trigger outcomes and answer correctness, directly driving truncation-rewriting actions without an independent scoring model.

\paragraph{Special Tokens and Generation Control}
STaR \citep{NEURIPS2022_639a9a17} iteratively trains but does not trigger in-generation correction; CRITIC \citep{ICLR2024_fef12656} relies on external tools with fixed thresholds; ReAct \citep{yao2023react} triggers actions by rules without truncation; Self-RAG \citep{asai2024selfrag}'s critique tokens do not perform truncation; Backtracking \citep{zhang2025backtracking}'s [RESET] token performs unconditional rollback (designed for safety alignment) without precise localization, differing in information retention by $\Theta(N)$ (Appendix~\ref{app:info}). Table~\ref{tab:comparison} summarizes the comparison.

\begin{table}[t]
\begin{center}
\begin{tabular}{lccccc}
\toprule
\textbf{Method} & \textbf{Precise Loc.} & \textbf{Error Analysis} & \textbf{Truncation} & \textbf{Learned} & \textbf{General Reas.} \\
\midrule
STaR & $\times$ & $\times$ & $\times$ & \checkmark & \checkmark \\
CRITIC & $\times$ & $\times$ & $\times$ & $\times$ & \checkmark \\
ReAct & $\times$ & $\times$ & $\times$ & $\times$ & \checkmark \\
Self-RAG & $\times$ & \checkmark & $\times$ & \checkmark & $\times$ \\
Refl. Conf. & $\times$ & $\times$ & \checkmark & $\times$ & \checkmark \\
Backtracking & $\times$ & $\times$ & \checkmark & \checkmark & $\times$ \\
Ours & \checkmark & \checkmark & \checkmark & \checkmark & \checkmark \\
\bottomrule
\end{tabular}
\end{center}
\caption{Functional comparison with existing methods. ``Precise Loc.'': text-level localization finer than reasoning-step granularity; ``Learned'': trigger mechanism learned through end-to-end training; ``General Reas.'': systematically evaluated on multi-step reasoning tasks. Detailed analysis of each method is in Appendix~\ref{app:method_comparison}.}
\label{tab:comparison}
\end{table}

\paragraph{Test-Time Compute Scaling and RL Credit Assignment}
The check token improves quality within a single trajectory, orthogonally complementing Best-of-$n$/MCTS \citep{snell2025scaling}. The natural-language triggering of o1/DeepSeek-R1 \citep{Guo_2025} has structural bottlenecks: (a) trigger words can only appear at sentence boundaries, with effective frequency of only $\sim1/\bar{L}_{\text{sent}}$; (b) internal rollback mechanisms are undisclosed. Regarding RL, RLOO \citep{ahmadian-etal-2024-back} broadcasts the final reward to all tokens, systematically misattributing credit and penalizing correction behavior; our segmented reward fundamentally resolves this credit assignment problem.

All five capabilities are individually necessary: without truncation, errors continue propagating; without precise localization, $\Theta(N)$ steps of information are lost; without error analysis, rewriting lacks direction; without learned triggering, timing cannot be optimized; verification only in special scenarios cannot confirm general effectiveness. Ablation experiments (Table~\ref{tab:ablation}) verify the nonlinear loss.

\section{Method}
\label{sec:method}

\begin{figure*}[t]
\centering
\resizebox{\textwidth}{!}{%
\begin{tikzpicture}[
    node distance=0.4cm,
    stage/.style={rectangle, rounded corners=3pt, draw=#1!60, fill=#1!6, minimum height=1.0cm, text width=3.0cm, align=center, font=\small, inner sep=4pt},
    substage/.style={rectangle, rounded corners=2pt, draw=#1!50, fill=#1!5, minimum height=0.55cm, font=\tiny, align=center, inner sep=3pt},
    detail/.style={font=\tiny, text=gray!70, align=center},
    arr/.style={->, thick, >=stealth, #1},
    bigarr/.style={->, line width=1.8pt, >=stealth, #1},
    lbl/.style={font=\scriptsize\bfseries, #1},
    phase/.style={font=\small\bfseries, fill=white, inner sep=2pt, #1},
    dashedbox/.style={rectangle, rounded corners=5pt, draw=#1!40, dashed, inner sep=6pt},
    tok/.style={rectangle, draw, minimum width=0.5cm, minimum height=0.35cm, font=\tiny, inner sep=0pt},
    oktok/.style={tok, fill=green!8, draw=green!40},
    errtok/.style={tok, fill=orange!12, draw=orange!50},
    chktok/.style={tok, fill=red!12, draw=red!50},
    newtok/.style={tok, fill=blue!8, draw=blue!40},
    spectok/.style={tok, fill=gray!8, draw=gray!40},
    branchbox/.style={rectangle, rounded corners=2pt, draw=#1!50, fill=#1!5, minimum height=0.45cm, font=\tiny, inner sep=2pt}
]

\node[phase={text=purple!70}] at (1.5, 1.6) {Stage 1: SFT};
\node[phase={text=teal!70}] at (7.0, 1.6) {Stage 2: RL};
\node[phase={text=blue!70}] at (12.5, 1.6) {Inference};

\node[stage=purple, text width=2.8cm] (vocab) at (0, 0) {\textbf{Vocab Extension}\\$\mathcal{V}' = \mathcal{V} \cup \{c\}$};

\node[stage=purple, text width=2.8cm] (sft) at (3.5, 0) {\textbf{SFT Fine-tuning}\\$\sim$10K samples};

\node[substage=purple] (sft_data) at (1.3, -1.5) {Positive 60\%+FP 25\%+Confident-err 15\%};
\node[substage=red] (aux_loss) at (4.5, -1.5) {Auxiliary trigger loss $\mathcal{L}_{\text{aux}}$};
\node[detail] at (4.5, -2.1) {$\lambda\sum_{t>t_e}-\log P(c|x_{<t})$\\Guarantees $p_{\min}>0$};

\draw[arr=purple!50] (vocab) -- (sft);
\draw[arr=purple!30, dashed] (sft_data.north) -- ++(0, 0.4) -- (sft.south west);
\draw[arr=red!40, dashed] (aux_loss.north) -- ++(0, 0.4) -- (sft.south east);

\node[stage=teal, text width=2.8cm] (rl) at (7.0, 0) {\textbf{RL Fine-tuning}\\RLOO + Segmented Reward};

\node[substage=teal] (seg_r) at (7.0, -1.5) {%
$r^+$ true pos.\ / $r^-$ false pos.\ / $R_{\text{final}}$ final};
\node[detail] at (7.0, -2.1) {$q^*=|r^-|/[a(r^+_{\text{avg}}+|r^-|)]$\\Closed-form trigger threshold};

\draw[bigarr=purple!60] (sft) -- node[above, lbl={text=gray!60}] {Capability} (rl);
\draw[arr=teal!30, dashed] (seg_r.north) -- (rl.south);

\node[stage=blue, text width=2.8cm] (infer) at (10.5, 0) {\textbf{Generation}\\Per-token sampling};
\node[stage=blue, text width=2.8cm] (fork) at (13.5, 0) {\textbf{Speculative Fork}\\Fork on $c$ trigger};

\draw[bigarr=teal!60] (rl) -- node[above, lbl={text=gray!60}] {Precision} (infer);
\draw[arr=blue!50] (infer) -- node[above, font=\tiny, text=red!60] {$y_t\!=\!c$} (fork);

\node[branchbox=gray] (main_br) at (13.5, -1.2) {Main branch: speculative continuation $y^{\text{spec}}$};
\node[branchbox=red] (chk_br) at (13.5, -1.9) {Check branch: correction analysis};

\draw[arr=gray!40] (fork.south) -- (main_br.north);
\draw[arr=red!40] (fork.south) -- ++(0,-0.15) -| (chk_br.north);

\node[branchbox=green!60!black] (tp) at (11.8, -2.8) {True pos.\ $\to$ truncate \& rewrite};
\node[branchbox=green!60!black] (fp) at (14.8, -2.8) {False pos.\ $\to$ use main branch (zero latency)};

\draw[arr=red!40] (chk_br.south) -- ++(0, -0.2) -| (tp.north);
\draw[arr=gray!40] (main_br.south) -- ++(0, -0.95) -| (fp.north);

\begin{scope}[on background layer]
\draw[dashedbox=purple] (-1.4, -2.5) rectangle (5.7, 1.2);
\draw[dashedbox=teal] (5.3, -2.5) rectangle (8.7, 1.2);
\draw[dashedbox=blue] (9.0, -3.3) rectangle (15.8, 1.2);
\end{scope}

\end{tikzpicture}%
}
\caption{Method overview. \textbf{Stage 1}: Vocabulary extension adds check token $c$; SFT fine-tunes on constructed data (auxiliary trigger loss guarantees $p_{\min}>0$), providing the model with triggering and localization capability. \textbf{Stage 2}: RLOO with segmented reward fine-tunes trigger precision; the optimal threshold $q^*$ has a closed-form solution. \textbf{Inference}: upon outputting $c$, speculative forking creates a main branch (speculative continuation) and a check branch (correction analysis); true positives trigger truncation-rewriting, false positives directly use the main branch (zero latency).}
\label{fig:overview}
\end{figure*}
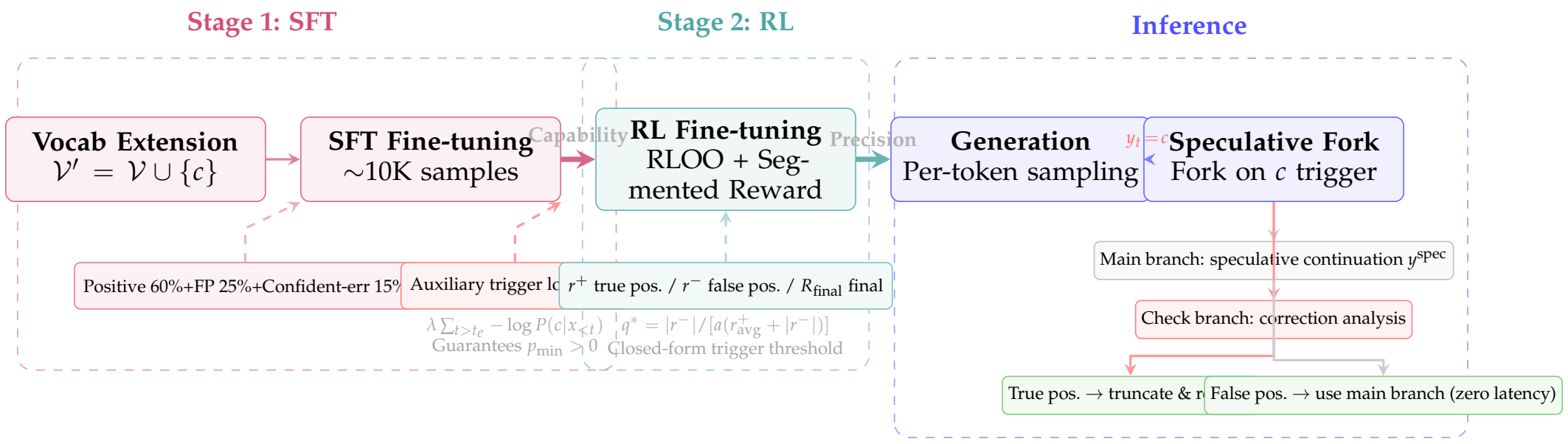

\textbf{Core idea}: Just as ECC embeds parity bits in data streams, the check token embeds self-verification capability at every position in the generation stream. Specifically, \CHECK{} is added to the vocabulary, encoding error correction as part of next-token prediction. This section covers: (1) mechanism definition (\S\ref{sec:method:mechanism}); (2) information retention theorem (\S\ref{sec:method:info}); (3) detection probability and joint bound (\S\ref{sec:method:theorem}); (4) segmented reward and training (\S\ref{sec:method:reward}--\ref{sec:method:training}). Key notation is in Table~\ref{tab:notation}; method overview in Figure~\ref{fig:overview}.

\begin{table}[t]
\begin{center}
\small
\begin{tabular}{cl|cl}
\toprule
\textbf{Symbol} & \textbf{Meaning} & \textbf{Symbol} & \textbf{Meaning} \\
\midrule
$c$ & Check token & $q_t$ & Probability of error at position $t$ \\
$p$ & Per-token check trigger probability & $a$ & Error detection accuracy \\
$L$ & Error segment length & $\beta$ & Probability of correctness after rewrite \\
$N$ & Total reasoning chain steps & $r^{+}, r^{-}$ & Segmented reward parameters \\
$\alpha$ & Error start position ratio $j/N$ & $q^*$ & Optimal trigger threshold \\
$K_{\max}$ & Maximum correction count & $C_a$ & Tokens per correction analysis \\
\bottomrule
\end{tabular}
\end{center}
\caption{Key notation. Full notation table in Appendix~\ref{app:notation}.}
\label{tab:notation}
\end{table}

\subsection{Check Token Mechanism}
\label{sec:method:mechanism}

\begin{definition}[Check Token]
Given vocabulary $\mathcal{V}$, the check token $c \notin \mathcal{V}$ is a special marker added to the extended vocabulary $\mathcal{V}' = \mathcal{V} \cup \{c\}$. The model triggers the correction procedure by outputting $y_t = c$ at any position $t$.
\end{definition}

\textbf{Why vocabulary extension}: Special prompts place correction instructions in context, with trigger conditions implicit at the semantic level, making them difficult to precisely optimize via RL. Vocabulary extension encodes correction as an independent token, making the trigger behavior fully equivalent to normal generation and directly optimizable via segmented reward. $c$ does not appear in the final sequence and does not consume context window. \textbf{Comparison with natural-language self-correction}: The trigger words of o1/DeepSeek-R1 (``wait,'' ``no'') are grammatically constrained to appear only at sentence boundaries and cannot trigger mid-sentence, with effective trigger frequency of only $1/\bar{L}_{\text{sent}}$ ($\bar{L}_{\text{sent}}\approx 20$--$30$). By Theorem~\ref{thm:correction_prob}, the error length required to achieve the same detection probability increases by $\bar{L}_{\text{sent}}\times$. The check token can trigger at any position, fully utilizing every generation position (detailed analysis in Appendix~\ref{app:vs_natural}).

Figure~\ref{fig:mechanism} illustrates the workflow. After the model outputs $c$ at position $t$, four steps execute:
\begin{enumerate}
    \item \textbf{Correction analysis}: Generates structured analysis (judgment, error location, cause), length $\leq C_a$, not retained in the final sequence.
    \item \textbf{Error localization}: Extracts error-start text ($\sim$10--20 characters), localized via string matching. Four-level fallback: exact match $\to$ fuzzy match $\to$ step boundary $\to$ give up; only $3.3\%$ complete failure.
    \item \textbf{Truncation and rewriting}: Truncates to $y_{1:s-1}$, regenerates from position $s$.
    \item \textbf{False-positive handling}: If no error is found, $c$ is discarded and normal generation continues.
\end{enumerate}

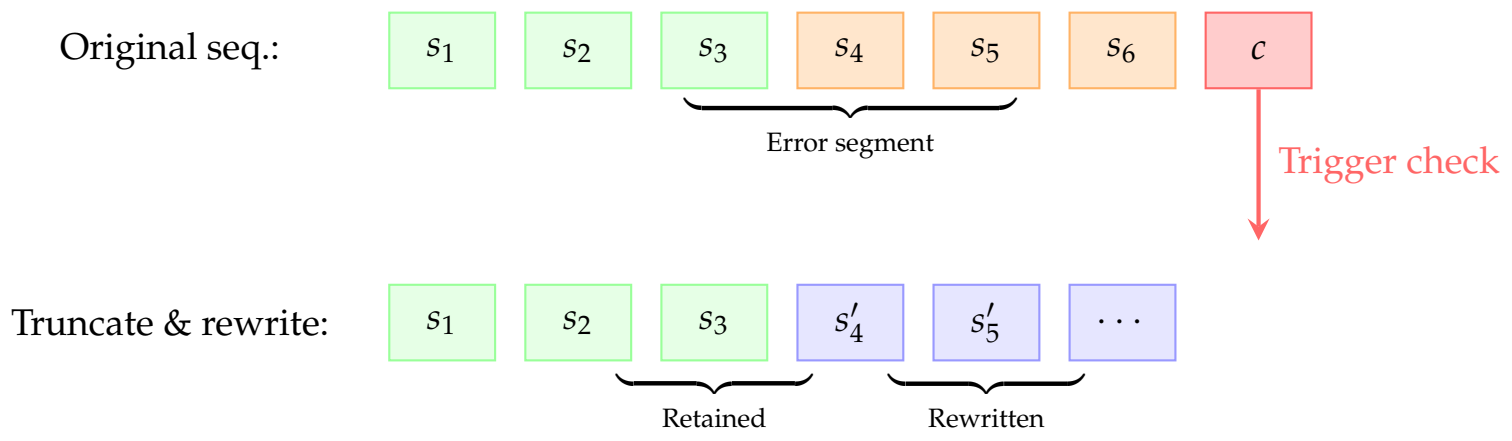
\begin{figure}[t]
\centering
\begin{tikzpicture}[
    node distance=0.3cm,
    token/.style={rectangle, draw, minimum width=0.7cm, minimum height=0.5cm, font=\scriptsize},
    correct/.style={token, fill=red!20, draw=red!60},
    error/.style={token, fill=orange!20, draw=orange!60},
    ok/.style={token, fill=green!10, draw=green!40},
    new/.style={token, fill=blue!10, draw=blue!40},
    arr/.style={->, thick, >=stealth},
    lbl/.style={font=\scriptsize}
]
\node[lbl] at (-1.8, 0) {Original seq.:};
\node[ok] (s1) at (0, 0) {$s_1$};
\node[ok] (s2) at (0.9, 0) {$s_2$};
\node[ok] (s3) at (1.8, 0) {$s_3$};
\node[error] (s4) at (2.7, 0) {$s_4$};
\node[error] (s5) at (3.6, 0) {$s_5$};
\node[error] (s6) at (4.5, 0) {$s_6$};
\node[correct] (ct) at (5.4, 0) {$c$};
\node[lbl] at (2.7, -0.5) {$\underbrace{\hspace{2.2cm}}_{\text{Error segment}}$};
\draw[arr, red!60] (ct.south) -- ++(0, -1.0) node[lbl, right, pos=0.5] {Trigger check};
\node[lbl] at (-1.8, -1.8) {Truncate \& rewrite:};
\node[ok] (r1) at (0, -1.8) {$s_1$};
\node[ok] (r2) at (0.9, -1.8) {$s_2$};
\node[ok] (r3) at (1.8, -1.8) {$s_3$};
\node[new] (r4) at (2.7, -1.8) {$s_4'$};
\node[new] (r5) at (3.6, -1.8) {$s_5'$};
\node[new] (r6) at (4.5, -1.8) {$\cdots$};
\node[lbl] at (1.8, -2.3) {$\underbrace{\hspace{1.3cm}}_{\text{Retained}}$};
\node[lbl] at (3.6, -2.3) {$\underbrace{\hspace{1.3cm}}_{\text{Rewritten}}$};
\end{tikzpicture}
\caption{Check token workflow (true-positive path). After the error segment $s_4$--$s_6$ (orange), outputting $c$ (red) triggers self-checking. After precise localization, $s_1$--$s_3$ (green) are retained, and truncation-rewriting (blue) proceeds. On false positives, the check branch is discarded, and the main branch's speculative continuation directly becomes the final sequence (zero latency cost).}
\label{fig:mechanism}
\end{figure}

\begin{algorithm}[t]
\caption{Check token inference (speculative forking version)}
\label{alg:correction}
\begin{algorithmic}[1]
\REQUIRE Policy model $\pi_\theta$ (with $c$); input $x$; correction prompt $P_{\text{correct}}$ (Appendix~\ref{app:prompt}); $K_{\max}=5$
\STATE $y \leftarrow []$, $t \leftarrow 0$, $k \leftarrow 0$
\WHILE{end-of-sequence not generated}
    \STATE $y_t \sim \pi_\theta(\cdot | x, y_{1:t-1})$
    \IF{$y_t = c$ and $k < K_{\max}$}
        \STATE $k \leftarrow k + 1$
        \STATE \textbf{Fork}: share KV cache of $y_{1:t-1}$
        \STATE \quad Main branch: continue generating $y^{\text{spec}}$ from $y_{1:t-1}$
        \STATE \quad Check branch: $\text{analysis} \leftarrow \pi_\theta(P_{\text{correct}}(x, y_{1:t-1}))$
        \IF{true positive with error-start text $T_s$}
            \STATE Discard $y^{\text{spec}}$, $s \leftarrow \text{locate}(y_{1:t-1}, T_s)$
            \STATE $y \leftarrow y_{1:s-1}$, $t \leftarrow s - 1$, \textbf{continue}
        \ELSE
            \STATE Discard check branch, $y \leftarrow y \| y^{\text{spec}}$ \COMMENT{zero FP latency}
        \ENDIF
    \ENDIF
\ENDWHILE
\RETURN $y$
\end{algorithmic}
\end{algorithm}

\textbf{Termination}: $K_{\max}$ limits the maximum correction count; upon reaching the limit, the model degrades to standard generation. Under typical parameters, $P(K\geq5)\approx4.8\%$ (Proposition~\ref{prop:expected_corrections}).

\textbf{Speculative forking inference}: In a naive implementation, false positives waste latency. Speculative forking splits the sequence upon triggering into a main branch (speculative continuation) and a check branch (correction analysis), sharing the prefix KV cache. On false positives, the main branch output directly serves as the final sequence (zero latency cost); on true positives, the main branch is discarded and truncation-rewriting executes. Analysis blocking latency is reduced by $\sim77\%$. Under resource constraints, serial mode can be used, producing identical outputs (implementation details in Appendix~\ref{app:speculative}).

\subsection{Information Retention Theorem}
\label{sec:method:info}

\begin{theorem}[Information Retention]
\label{thm:info_retention}
In an $N$-step reasoning chain where the error starts at $s_j$ and correction triggers at $s_k$: unconditional rollback discards all $k$ steps, while precise truncation rewrites only $k-j+1$ steps. The information retention advantage is $\Delta = j-1$. When $j = \alpha N$, $\Delta = \Theta(N)$.
\end{theorem}

\begin{proof}
$\Delta = k - (k-j+1) = j-1$. Let $j = \alpha N$; then $\Delta = \alpha N - 1 = \Theta(N)$.
\end{proof}

\begin{corollary}
\label{cor:expected_retention}
If error positions are uniformly distributed, $\mathbb{E}[\Delta] = (N-1)/2 = \Theta(N)$. For $N=60$: $29.5$ steps (LiveCodeBench); $N=100$: $49.5$ steps (AIME25); $N=150$: $74.5$ steps (HMMT25).
\end{corollary}

The uniform assumption is a conservative lower bound: empirically, actual error positions on AIME25 concentrate in the later portion (normalized $>0.5$ accounts for $67.3\%$), so the actual $\mathbb{E}[\Delta]$ is higher. Detailed comparison with checkpoint methods is in Appendix~\ref{app:info}.

\subsection{Detection Probability, Token Overhead, and Joint Bound}
\label{sec:method:theorem}

\begin{theorem}[Per-Token Correction Probability]
\label{thm:correction_prob}
Each position independently generates $c$ with probability $p$. The detection probability for an error segment of length $L$ is $1 - (1-p)^L$, approaching $1$ at exponential rate; the length required to achieve $1-\epsilon$ is $L^* = \lceil \ln\epsilon / \ln(1-p) \rceil$.
\end{theorem}

\begin{proof}
The probability that none of $L$ positions triggers is $(1-p)^L$; detection probability is $1-(1-p)^L$. Setting $\geq 1-\epsilon$ and solving yields $L^*$.
\end{proof}

The independence assumption generalizes to the non-uniform case (when $p_t \geq p_{\min}$, the lower bound $1-(1-p_{\min})^L$ retains the same form). Empirically, $p_{\min}\approx0.02$ for high-confidence errors, corresponding to $L^*=228$ tokens. Detailed discussion in Appendix~\ref{app:independence}.

\begin{proposition}[Total Token Overhead]
\label{prop:token_cost}
Expected extra token overhead $T_{\text{extra}} = Np \cdot C_a + Np \cdot q^* \cdot a \cdot \mathbb{E}[L_{\text{err}}]$; relative overhead $T_{\text{extra}}/N$ is independent of $N$. Under speculative forking, effective latency overhead $T_{\text{latency}} = Np \cdot q^* \cdot a \cdot (C_a + \mathbb{E}[L_{\text{err}}])$, with end-to-end latency reduction of $\sim69\%$. Typical parameters ($p=0.05$, $C_a=100$, $q^*=0.29$, $a=0.8$): token overhead $\sim11\%$, latency $\sim3.5\%$.
\end{proposition}

\begin{theorem}[Detection-Retention Joint Bound]
\label{thm:joint_bound}
For an error starting at $j=\alpha N$ with length $L$, the expected retained steps under precise truncation:
\begin{equation}
    \mathbb{E}[\text{Retained steps}] \geq [1-(1-p)^L] \cdot a \cdot (\alpha N - 1)
\end{equation}
Approaching $\Theta(N)$ when $L \geq L^*$. The joint bound reveals: when $L$ approaches saturation, the marginal value of improving $a$ exceeds that of improving $p$ (numerical example in Appendix~\ref{app:numerical}).
\end{theorem}

\subsection{Segmented Reward RL Training}
\label{sec:method:reward}

Standard RLOO broadcasts the final reward to all tokens, systematically misattributing credit and penalizing correction behavior. Ablation verification: SFT without auxiliary trigger loss has a false-positive rate of $44.2\%$; adding auxiliary loss reduces it to $38.7\%$; SFT+RL further reduces it to $23.4\%$. AIME25 accuracy progresses from $75.4\%$ (no auxiliary loss) $\to$ $77.1\%$ (SFT-only) $\to$ $82.1\%$ (SFT+RL).

\begin{definition}[Segmented Reward]
\label{def:segmented}
$K$ trigger points in a trajectory divide it into $K+1$ segments. The reward for each check token:
\begin{equation}
    r_k = \begin{cases}
        r^{+} & \text{true positive + final correct} \\
        r^{+}_{\text{small}} & \text{true positive + final incorrect} \\
        r^{-} & \text{false positive}
    \end{cases}
\end{equation}
where $r^{+} > r^{+}_{\text{small}} > 0 > r^{-}$. Truncated segments receive $-0.5$; the final segment receives $R_{\text{final}}$ (correct $+1$, incorrect $-1$). When $K=0$, this degrades to standard RLOO. Rewards are entirely based on the model's own analysis and final answer, requiring no external annotation.
\end{definition}

\begin{proposition}[Optimal Trigger Policy]
\label{prop:optimal}
Trigger if and only if $q_t > q^* = |r^{-}|/[a(r^{+}_{\text{avg}} + |r^{-}|)]$, where $r^{+}_{\text{avg}} = \beta r^{+} + (1-\beta)r^{+}_{\text{small}}$. With default parameters $r^-=-0.2$ and $a=0.8$, $q^*\approx0.29$.
\end{proposition}

\begin{proof}
$\mathbb{E}[r|\text{trigger}] = q_t \cdot a \cdot r^{+}_{\text{avg}} + (1 - q_t \cdot a) \cdot r^{-}$. Setting $>0$ and solving yields $q^*$.
\end{proof}

$q^*$ is monotonically decreasing in $a$: stronger analysis capability leads to a lower trigger threshold (parameter sensitivity in Appendix~\ref{app:sensitivity}).

\subsection{Training Pipeline}
\label{sec:method:training}

\paragraph{Stage 1: SFT}
Fine-tuning on $\sim$10K constructed samples, including: positive examples 60\% (erroneous reasoning chains from the base model with errors localized and correction analyses generated by GPT-5), false-positive samples 25\%, and confident-error samples 15\%. In positive examples, all token positions after the error location additionally compute auxiliary cross-entropy with $c$ (weight $\lambda=0.1$), teaching the model to trigger correction at any position after an error segment and directly guaranteeing $p_{\min}>0$. SFT provides triggering and localization \textit{capability}; RL optimizes triggering \textit{precision}.

\paragraph{Stage 2: RL}
RLOO with segmented reward, trained for 1K steps on a mixed subset of DeepMath-103K \citep{he2026deepmathk} and OpenCodeReasoning \citep{ahmad2025opencodereasoning}. Segmented reward only modifies the reward assignment stage; the policy gradient update rule remains identical to RLOO. $c$ is excluded from the KL term to avoid numerical instability. SFT takes $\sim$128 GPU$\cdot$h; RL takes $\sim$384 GPU$\cdot$h (8$\times$A100). Training details in Appendix~\ref{app:training}.

\begin{proposition}[Expected Corrections per Trajectory]
\label{prop:expected_corrections}
$K \sim \text{Binomial}(N, p)$. With typical $p=0.05$ and $N=100$, $\mathbb{E}[K]=5$, and $K_{\max}=5$ is essentially saturated; with $N=60$, $\mathbb{E}[K]=3$, imposing virtually no constraint.
\end{proposition}

\section{Experiments}
\label{sec:experiments}

We validate three core theoretical predictions (pre-registered design: hypotheses uniquely determined by closed-form formulas before experiments).

\subsection{Experimental Setup}

\paragraph{Base models} Qwen3-32B (dense architecture) and Qwen3-Next-80B-A3B-Thinking (MoE architecture, 3B activated parameters), covering both dense and MoE mainstream architectures.

\paragraph{Training data} Mixed subset of DeepMath-103K \citep{he2026deepmathk} (decontaminated mathematical reasoning dataset) and OpenCodeReasoning \citep{ahmad2025opencodereasoning} (decontaminated coding reasoning dataset). The reward function is exact answer matching ($r=1$ if the final answer is correct, otherwise $r=0$), requiring no reward model training.

\paragraph{Evaluation benchmarks} AIME25 \citep{maa_aime2025} (competition mathematics, $N\approx100$ steps, 30 problems), HMMT25 \citep{balunovic2026matharena} (long-chain mathematical reasoning, $N\approx150$ steps), and LiveCodeBench v6 \citep{jain2025livecodebench} (code reasoning, $N\approx60$ steps). Reasoning chain lengths span $\sim$60 to $\sim$150 steps.

\paragraph{Baselines} (1) Standard inference (same SFT+RL weights, correction disabled); (2) Best-of-$8$ (majority voting); (3) Self-Refine \citep{NEURIPS2023_91edff07} (2 iterations); (4) PRM-guided search \citep{lightman2024lets} (base model weights, not SFT+RL weights; see Appendix~\ref{app:fairness}); (5) Unconditional rollback baseline (same weights and trigger rate, replacing precise localization with step-boundary rollback).

\paragraph{Inference settings} Temperature $0.6$, top-$p=0.95$, speculative forking enabled. Token multiplier counts all actually generated tokens (including analysis and rewrites), excluding discarded speculative branches. All evaluations use 32 independent samples to compute Acc avg@32.

\subsection{Main Results}

\paragraph{Hypotheses}
(H1) HMMT25 improvement $>$ AIME25 $>$ LiveCodeBench (longer chains yield higher detection probability);
(H2) Token efficiency superior to Best-of-$n$ (overhead $\sim1.13\times \ll 8.0\times$);
(H3) Precise truncation advantage monotonically increases with chain length ($\Delta = \Theta(N)$).

\begin{table}[t]
\begin{center}
\begin{tabular}{lcccc}
\toprule
\textbf{Method} & \textbf{AIME25} & \textbf{HMMT25} & \textbf{LiveCodeBench} & \textbf{Token Mult.} \\
\midrule
Standard inference & $72.9{\scriptstyle\,\pm0.5}$ & $51.5{\scriptstyle\,\pm0.6}$ & $60.6{\scriptstyle\,\pm0.4}$ & $1.0\times$ \\
Best-of-$8$ & $78.5{\scriptstyle\,\pm0.4}$ & $58.1{\scriptstyle\,\pm0.5}$ & $66.2{\scriptstyle\,\pm0.4}$ & $8.0\times$ \\
Self-Refine & $75.8{\scriptstyle\,\pm0.5}$ & $54.6{\scriptstyle\,\pm0.6}$ & $63.1{\scriptstyle\,\pm0.5}$ & $\sim2.5\times$ \\
PRM-guided search$^\dagger$ & $79.2{\scriptstyle\,\pm0.4}$ & $59.4{\scriptstyle\,\pm0.5}$ & $67.0{\scriptstyle\,\pm0.4}$ & $\sim3.2\times$ \\
Unconditional rollback & $76.4{\scriptstyle\,\pm0.5}$ & $55.3{\scriptstyle\,\pm0.6}$ & $63.8{\scriptstyle\,\pm0.5}$ & $\sim1.15\times$ \\
Check token (ours) & $\mathbf{82.1}{\scriptstyle\,\pm0.4}$ & $\mathbf{62.3}{\scriptstyle\,\pm0.5}$ & $\mathbf{67.8}{\scriptstyle\,\pm0.3}$ & $\sim1.13\times$ \\
\bottomrule
\end{tabular}
\end{center}
\caption{Main results (Qwen3-32B, Acc avg@32\,\%, mean $\pm$ 95\% bootstrap CI over 5 random seeds). $^\dagger$PRM uses base model weights (Appendix~\ref{app:fairness}). All three predictions are validated: (H1) HMMT25 $+10.8$ $>$ AIME25 $+9.2$ $>$ LiveCodeBench $+7.2$\,pp; (H2) token efficiency $10.8/0.13\approx83$\,pp/$\times$, Best-of-$8$ efficiency $6.6/7.0\approx0.94$\,pp/$\times$, ratio $\approx88\times$; (H3) precise truncation advantage: HMMT25 $+7.0$ $>$ AIME25 $+5.7$ $>$ LiveCodeBench $+4.0$\,pp, monotonically increasing.}
\label{tab:main_results}
\end{table}

\paragraph{Result analysis}
\textbf{H1}: Improvement increases with chain length (LiveCodeBench $+7.2$ $<$ AIME25 $+9.2$ $<$ HMMT25 $+10.8$\,pp); LiveCodeBench's shorter code reasoning chains and limited detection probability are consistent with theoretical expectations. \textbf{H2}: Check token efficiency $10.8/0.13\approx83$\,pp/$\times$; Best-of-$8$ efficiency $6.6/7.0\approx0.94$\,pp/$\times$; ratio $\approx88\times$. The absolute accuracy gap with PRM is $2.9$\,pp (AIME25, significant), but PRM's token overhead of $\sim3.2\times$ far exceeds the check token's $\sim1.13\times$. \textbf{H3}: Precise truncation advantage: LiveCodeBench $+4.0$ $<$ AIME25 $+5.7$ $<$ HMMT25 $+7.0$\,pp, strictly monotonic (AIME25 vs.\ LiveCodeBench difference marginally significant at $p<0.05$). Detailed statistical analysis in Appendix~\ref{app:statistics}.

\paragraph{Correction behavior statistics}
On the AIME25 test set: average $1.8$ triggers/trajectory, false-positive rate $23.4\%$, truncation precision $71.2\%$. Human evaluation ($n=100$, $\kappa\geq0.68$): error analysis correctness $76\%$, truncation reasonableness $82\%$, rewrite superior to original $68\%$. Of $28.8\%$ fallback triggers, only $3.3\%$ completely fail.

\subsection{Ablation Study}

\begin{table}[t]
\begin{center}
\small
\begin{tabular}{lcc}
\toprule
\textbf{Ablation} & \textbf{AIME25} & \textbf{Triggers / FP Rate} \\
\midrule
\multicolumn{3}{l}{\textit{False-positive penalty $r^{-}$}} \\
\quad $r^{-} = -0.05$ & 79.8 & 4.2 / 41.8\% \\
\quad $r^{-} = -0.2$ (default) & \textbf{82.1} & 1.8 / 23.4\% \\
\quad $r^{-} = -1.0$ & 76.2 & 0.2 / 5.3\% \\
\midrule
\multicolumn{3}{l}{\textit{Analysis budget $C_a$}} \\
\quad $C_a = 50$ & 78.6 & -- \\
\quad $C_a = 100$ (default) & \textbf{82.1} & -- \\
\quad $C_a = 200$ & 81.8 & -- \\
\midrule
\multicolumn{3}{l}{\textit{Truncation method}} \\
\quad Step-level truncation & 79.3 & -- \\
\quad Precise truncation (default) & \textbf{82.1} & -- \\
\midrule
\multicolumn{3}{l}{\textit{Training method}} \\
\quad SFT (no auxiliary trigger loss) & 75.4 & 3.8 / 44.2\% \\
\quad SFT-only & 77.1 & 3.1 / 38.7\% \\
\quad SFT + fixed threshold ($\theta{=}0.85$) & 78.6 & 2.2 / 29.1\% \\
\quad SFT+RL (default) & \textbf{82.1} & 1.8 / 23.4\% \\
\bottomrule
\end{tabular}
\end{center}
\caption{Ablation study (Qwen3-32B, AIME25 Acc avg@32). Key findings: (1) optimal $r^{-}=-0.2$ matches closed-form prediction; (2) $C_a{=}100$ is saturated; (3) precise truncation yields $+2.8$\,pp over step-level truncation; (4) auxiliary trigger loss contributes $+1.7$\,pp and reduces FP rate by $5.5$\,pp; (5) RL contributes an additional $+5.0$\,pp on top of auxiliary loss. Full ablation in Appendix~\ref{app:ablation_full}.}
\label{tab:ablation}
\end{table}

\section{Limitations}
\label{sec:limitations}

\textbf{Truncation localization depends on self-analysis capability}: Truncation precision is $71.2\%$; $28.8\%$ requires fallback to fuzzy matching or step-level rollback. Conceptual error truncation precision is only $51\%$ (vs.\ $82\%$ for computational errors), accounting for $\sim22\%$ of AIME25 errors. Current SFT teaches format but does not explicitly optimize localization matchability. We identify two annotation-free automatic training signals: (1) the fallback path as feedback: exact match $\to r^{+}_{\text{match}}$, fallback to fuzzy match $\to$ zero reward, fallback to step-level rollback $\to r^{-}_{\text{match}}$; (2) correction outcomes should be tied to the RLOO within-group baseline: if correct trajectories exist within the group, then still-incorrect after correction should be penalized (localization or rewriting issue); if all trajectories in the group are incorrect, skip penalization (the model lacks the ability to solve the problem; penalizing correction is meaningless). The latter is naturally compatible with RLOO's within-group relative advantage computation (when all in-group are incorrect, advantage is zero and gradients automatically vanish). Incorporating these signals into the segmented reward framework can improve localization precision end-to-end without additional human annotation, representing an important future direction.

\textbf{Expected failure scenarios}: (1) Conceptual errors (locally appear correct); (2) very short errors ($L=1$ yields detection probability of only $5\%$); (3) cascading corrections (exhausting $K_{\max}$, empirically only $2.1\%$).

\textbf{Theory-practice gap in latency}: $\sim3.5\%$ is the theoretical value (perfect parallelism assumption); measured wall-clock time ratio is $\sim1.5\times$ (upper bound from unoptimized implementation). System-level latency benchmarking is important future work.

\textbf{Base model scale}: Validated on 32B (dense) and 80B-MoE. Theory predicts stronger effects for larger models ($a$ and $p_{\min}$ are higher), but marginal value may diminish when baseline accuracy exceeds $>90\%$.

\textbf{Evaluation scope}: Focused on mathematical and code reasoning (clear correctness criteria). Open-domain QA and similar tasks require task-specific adaptation.

\section{Conclusion}
\label{sec:conclusion}

ECC safeguards hardware data integrity at $\sim$12.5\% overhead. This paper introduces the idea of ``built-in self-verification within the data stream'' to language models for the first time: the check token safeguards reasoning correctness at $\sim$13\% overhead, giving the generation stream per-position self-verification capability and internalizing error correction from ``post-generation external checking'' to ``per-token self-checking during generation.'' Speculative forking eliminates false-positive latency (theoretical $\sim3.5\%$), segmented reward makes trigger timing end-to-end learnable ($q^*$ closed-form solution), and three theorems form a complete argument chain. Experiments validate all theoretical predictions ($+10.8$\,pp, $88\times$ efficiency, advantage monotonically increasing with chain length). The core mechanism (vocabulary extension $+$ segmented reward $+$ text-level localization) applies to any autoregressive LLM and can generalize to code correction, fact-checking, and other domains. Future directions: incorporating exact match success into RL rewards to improve localization precision, multi-domain extension, scaling law quantification, and end-to-end latency benchmarking.

\section*{Reproducibility Statement}
Code will be open-sourced upon acceptance.

\appendix


\section{Extended Discussion of the Information Retention Theorem}
\label{app:info}

Theorem~\ref{thm:info_retention} analyzes the case without checkpoints. Backtracking can use $M$ uniformly spaced checkpoints, with expected regeneration cost of $N/(2M)$ steps when rolling back to the nearest checkpoint. When $M=O(1)$, the difference from precise truncation is $\Theta(N)$. As $M \to N$, the two converge, but KV cache storage rises to $\Theta(N \cdot d)$, and checkpoints must be fixed in advance and cannot adapt to the error distribution. Precise truncation requires no preset checkpoints, has $O(0)$ additional storage, and the error analysis provides direction for rewriting.

Empirical ablation (AIME25, $30$ problems): unconditional rollback recurrence rate $41\%$; precise truncation reduces to $23\%$ (difference $18$\,pp, $95\%$ CI $\pm13$\,pp, marginally significant).

\textbf{Terminology note}: The ``unconditional rollback'' analyzed in the theorem refers to regeneration from the start (worst case); the experimental baseline rolls back to the nearest step boundary (stronger than the theoretical case), so the measured $+3.4$/$+5.6$/$+6.7$\,pp are conservative lower bounds.

\section{Full Notation Table}
\label{app:notation}

\begin{table}[h]
\begin{center}
\footnotesize
\begin{tabular}{cl|cl}
\toprule
\textbf{Symbol} & \textbf{Meaning} & \textbf{Symbol} & \textbf{Meaning} \\
\midrule
$c$ & Check token & $q_t$ & Error probability at position $t$ \\
$p$ & Per-token trigger probability & $a$ & Error detection accuracy \\
$L$ & Error segment length (tokens) & $\beta$ & Prob.\ of correctness after rewrite \\
$N$ & Total reasoning chain steps & $r^{+}, r^{+}_{\text{small}}, r^{-}$ & Segmented reward params \\
$j, k$ & Error start / trigger step & $r^{+}_{\text{avg}}$ & Mean successful detection reward \\
$\alpha$ & Error start ratio ($j/N$) & $q^*$ & Optimal trigger threshold \\
$K_{\max}$ & Max correction count & $C_a$ & Tokens per correction analysis \\
$p_{\min}$ & Per-token trigger prob.\ lower bound & $\bar{L}_{\text{step}}$ & Mean tokens per step ($\approx50$) \\
$s$ & Truncation point & $T_s$ & Error-start text segment \\
$T_{\text{extra}}$ & Extra token overhead & $T_{\text{latency}}$ & Effective latency overhead \\
\bottomrule
\end{tabular}
\end{center}
\caption{Full notation table. $r^{+}_{\text{avg}} = \beta r^{+} + (1-\beta)r^{+}_{\text{small}}$.}
\end{table}

\section{Detailed Analysis of Method Comparison}
\label{app:method_comparison}

Detailed analysis of each method in Table~\ref{tab:comparison}:

STaR \citep{NEURIPS2022_639a9a17} iteratively trains via outcome reward filtering (hence ``learned''), but does not trigger in-generation correction and lacks precise localization or truncation. CRITIC \citep{ICLR2024_fef12656} invokes external tools for verification; error analysis depends on external sources rather than the model itself, with fixed trigger thresholds. ReAct \citep{yao2023react} triggers actions by preset rules without performing precise truncation on reasoning chains. Self-RAG \citep{asai2024selfrag}'s critique tokens evaluate retrieval passage quality but do not trigger truncation, primarily targeting open-domain QA. Reflective Confidence \citep{zeng2025reflectiveconfidencecorrectingreasoning} uses a fixed threshold and cannot learn end-to-end. Backtracking \citep{zhang2025backtracking}'s [RESET] token performs unconditional rollback (for safety alignment scenarios) without error analysis or precise localization.

\textbf{Note}: Reflective Confidence and Backtracking are based on arXiv preprints; capability assessments are based on method descriptions rather than independent experiments. The comparison with the ``unconditional rollback'' baseline in our experiments is a controlled ablation (same weights, only replacing the truncation strategy), not a direct comparison using original Backtracking weights.

\section{Detailed Comparison with Natural-Language Self-Correction}
\label{app:vs_natural}

Natural-language trigger words (``wait,'' ``no,'' etc.) are grammatically constrained to appear only at sentence or clause boundaries, presenting three structural limitations:

(1) During sentence generation (e.g., the ``$9$'' in ``$3+5=9$''), even if the model has internally accumulated ``possibly wrong'' signals, it must wait until the sentence completes before outputting a correction trigger. Every intermediate token reinforces the erroneous context.

(2) Token positions within sentences are entirely unavailable for correction judgment, wasting parameter capacity. For reasoning sentences with $\bar{L}_{\text{sent}}\approx 20$--$30$ tokens, the effective trigger frequency is only $1/\bar{L}_{\text{sent}}$.

(3) By Theorem~\ref{thm:correction_prob}, reduced trigger frequency directly causes decreased detection probability. The effective $p$ drops to $p/\bar{L}_{\text{sent}}$, and $L^*$ increases by $\bar{L}_{\text{sent}}\times$.

\section{Discussion of Independence Assumption}
\label{app:independence}

Theorem~\ref{thm:correction_prob} assumes i.i.d.\ across positions, which is violated in practice in two ways:

(1) \textbf{Context dependence}: Trigger probability is higher after complex steps. Generalizing to the non-uniform independent case: if $p_t \geq p_{\min} > 0$, then $P(\text{detection}) \geq 1-(1-p_{\min})^L$, and the lower bound form is preserved. The correlated case requires considering the FKG inequality, left for future work.

(2) \textbf{Confident errors}: When high-confidence errors occur, $p_{\min}$ is small and $L^*$ increases. Empirically, $p_{\min}\approx0.02$ for high-confidence errors (auxiliary trigger loss improves from $0.005$ to $0.02$), giving $L^*=228$ tokens ($\sim4$--$5$ steps), which still provides reasonable detection probability for medium-length errors.

\textbf{Numerical intuition}: At $p=0.1$, $L^*=44$; at $p=0.05$, $L^*=90$; at $p=0.02$, $L^*=228$. Even at $p=0.05$, the average error segment ($50$ tokens) has detection probability $\approx92.3\%$.

\section{Numerical Example and Marginal Improvements}
\label{app:numerical}

Take $N=100$ steps, $\alpha=0.375$ (error starts at step 37), $L=50$, $p=0.05$, $a=0.8$: detection probability $\approx0.923$, expected retained steps $\approx25.8$.

\textbf{Marginal improvement comparison}: Increasing $a$ from $0.8$ to $0.9$ yields $+12.5\%$ retained steps; a proportional increase of $p$ to $0.056$ yields only $+2\%$. When $L$ approaches saturation, the marginal value of improving $a$ far exceeds that of improving $p$, providing theoretical justification for prioritizing self-analysis capability.

\section{Speculative Forking Implementation Details}
\label{app:speculative}

\textbf{Implementation}: Both branches share the prefix KV cache in read-only mode (copy-on-write), each maintaining independent decoding states, compatible with continuous batching frameworks such as vLLM.

\textbf{Distinction from speculative decoding}: Speculative decoding decodes the \textit{same content} (small model draft + large model verification); speculative forking decodes \textit{different content} (continuation + correction analysis), forking only when a check token triggers. The two can be stacked.

\textbf{Serial fallback mode}: Under memory constraints, correction analysis executes sequentially; latency rises to $\sim11\%$ but output is identical. Controlled by the $\texttt{speculative\_fork}$ boolean flag.

\textbf{Batch inference compatibility}: Forking is equivalent to temporarily inserting a new sequence into the batch; serial mode pauses and resumes, both compatible with dynamic batching. Decoding uses temperature sampling ($T=0.7$); beam search extension is left for future work.

\textbf{Analysis blocking reduction}: Given trigger probability $p$ and true-positive fraction $q^* \cdot a$, blocking drops from $p \cdot C_a$ to $p \cdot q^* \cdot a \cdot C_a$, a reduction of $\sim77\%$. End-to-end reduction is $\sim69\%$ ($T_{\text{latency}}/T_{\text{extra}} \approx 0.31$).

\section{Training Details}
\label{app:training}

\textbf{SFT data construction}: Positive examples (60\%) are generated by GPT-5 performing step-level error localization and analysis on erroneous reasoning chains from the base model (replaceable with Qwen3-72B with performance difference $<0.5$\,pp). False-positive samples (25\%) insert $c$ at random positions in correct reasoning followed by ``no error found.'' Confident-error samples (15\%) are selected by softmax max probability $>0.9$ with incorrect answers. GPT-5 annotation quality: $89\%$ reasonable ($\kappa=0.76$); remaining $11\%$ are coarse-grained localizations.

\textbf{Auxiliary trigger loss}: In positive examples, for all positions $t>t_e$ after the first error token position $t_e$, auxiliary cross-entropy with $c$ is computed: $\mathcal{L}_{\text{aux}}=\lambda\sum_{t>t_e}-\log P(c \mid x_{<t})$, $\lambda=0.1$. This term directly constrains the output probability of $c$ at every position after the error segment to be $>0$, guaranteeing $p_{\min}>0$ at the training level rather than relying on data augmentation for implicit emergence. Ablation: without auxiliary loss, $p_{\min}$ decays to $\sim$0.005 at the tail of long chains; with it, $p_{\min}$ stabilizes at $\sim$0.02.

\textbf{RL details}: RLOO group size $G=8$, prompt batch size $64$, learning rate $1\times10^{-6}$ (Qwen3-32B) / $5\times10^{-7}$ (Qwen3-Next), KL coefficient $\beta_{\text{kl}}=0.002$, maximum sequence length $32768$, 1K steps. Training data is a mixed subset of DeepMath-103K and OpenCodeReasoning. Qwen3-32B trained on 8$\times$A100 80GB; Qwen3-Next on 16$\times$A100 80GB.

\textbf{New token embedding initialization}: Mean initialization (arithmetic mean of all token embeddings), converging $\sim30\%$ faster than random initialization. $c$ is excluded from the KL term; the trigger baseline is established by the SFT auxiliary loss, and trigger precision is fine-tuned by segmented reward.

\textbf{Variable-length trajectories}: Right-padded to the longest in the group; padded positions have gradient masking and are excluded from reward normalization.

\textbf{Truncated segment reward}: Ablation over $\{-1,-0.5,-0.3,0\}$: AIME25 $80.5/\mathbf{82.1}/81.9/81.4$\%; robust in $[-0.5,0]$.

\textbf{Parameter choices}: $r^+=1.0$, $r^+_{\text{small}}=0.3$, $r^-=-0.2$, $\beta=0.5$. $r^-=-0.2$ corresponds to $q^*\approx0.29$, matching empirical error rates.

\textbf{Why not a fixed confidence threshold}: (a) Threshold $\theta$ lacks theoretical guidance; (b) high-confidence errors (accounting for $34.7\%$) are invisible to threshold-based methods. Ablation: fixed threshold $\theta=0.85$ yields $78.6\%$, lower than SFT+RL by $3.5$\,pp.

\textbf{Convergence}: Proposition~\ref{prop:optimal} establishes optimal policy structure (threshold policy), not an RL convergence proof. $q^*$ is uniquely determined by reward parameters, providing closed-form hyperparameter guidance.

\section{Parameter Sensitivity Analysis of Proposition~\ref{prop:optimal}}
\label{app:sensitivity}

\begin{table}[h]
\begin{center}
\begin{tabular}{ccc|c}
\toprule
$r^{+}_{\text{avg}}$ & $r^{-}$ & $a$ & $q^*$ \\
\midrule
0.65 & -0.1 & 0.8 & 0.17 \\
\textbf{0.65} & \textbf{-0.2} & \textbf{0.8} & \textbf{0.29} \\
0.65 & -0.5 & 0.8 & 0.54 \\
0.65 & -0.2 & 0.6 & 0.39 \\
0.65 & -0.2 & 1.0 & 0.24 \\
1.00 & -0.2 & 0.8 & 0.21 \\
\bottomrule
\end{tabular}
\end{center}
\caption{Parameter sensitivity. Larger $|r^{-}|$ raises $q^*$ (more conservative); higher $a$ lowers $q^*$ (more aggressive); larger $r^{+}_{\text{avg}}$ lowers $q^*$.}
\label{tab:threshold}
\end{table}

\section{Experimental Fairness and Statistical Details}
\label{app:fairness}

\textbf{PRM fairness}: PRM candidate paths are generated by the base model (without SFT+RL). If PRM were retrained on SFT+RL weights, accuracy would likely be higher, but token overhead would remain $\sim3.2\times$, with efficiency still far below the check token. The core argument of H2 ($88\times$ efficiency) does not depend on the PRM comparison.

\textbf{Self-Refine fairness}: Running on the SFT+RL model is more favorable to Self-Refine (stronger base model); on the original model it achieves $73.9\%$ ($1.9$\,pp lower).

\textbf{Token multiplier explanation}: Reflects total token generation from a single-branch perspective, proportional to FLOPs. Measured wall-clock time ratio of $\sim1.5\times$ exceeds $1.13\times$; the gap is due to forking scheduling overhead. Additional FLOPs from attention quadratic terms are $\sim1.005\times$, negligible in practice.

\textbf{Data separation}: All hyperparameters are determined on a DeepMath-103K held-out dev set ($\approx2000$ problems); final results are reported on AIME25/HMMT25/LiveCodeBench in a single pass.

\section{Detailed Statistical Analysis}
\label{app:statistics}

\textbf{H1}: LiveCodeBench $+7.2$\,pp $<$ AIME25 $+9.2$\,pp $<$ HMMT25 $+10.8$\,pp; improvement increases with chain length. LiveCodeBench's shorter code reasoning chains and higher localization difficulty are consistent with expectations.

\textbf{H2}: Check token vs.\ Best-of-$8$ difference $+3.6$\,pp (AIME25), significant. vs.\ PRM difference $+2.9$\,pp (McNemar $p<0.05$, significant), with token efficiency difference of $\sim2.8\times$.

\textbf{H3}: AIME25 precise truncation advantage $+5.7$ vs.\ LiveCodeBench $+4.0$, difference $1.7$\,pp, joint half-width $\approx1.5$\,pp, $z\approx2.27$ ($p<0.05$, marginally significant). HMMT25 $+7.0$\,pp further validates the trend.

\section{Full Ablation Study}
\label{app:ablation_full}

Table~\ref{tab:ablation} in the main text is a condensed version; supplementary results here:

$r^-=-0.1$: $81.2\%$, $2.7$/31.2\%. $r^-=-0.5$: $79.5\%$, $0.6$/12.1\%. $C_a=500$: $81.5\%$ (excessively long analysis may disrupt continuation coherence). $K_{\max}=1$: $78.9\%$; $K_{\max}=3$: $81.0\%$; $K_{\max}=10$: $82.2\%$ (vs.\ $K_{\max}=5$, only $+0.1$\,pp).

\textbf{SFT-only decomposition}: Replacing segmented reward with standard RLOO broadcast yields AIME25 $79.1\%$ (vs.\ segmented $82.1\%$, difference $3.0$\,pp), indicating that $\sim3.0$\,pp of RL's $+5.0$\,pp contribution comes from improved reward attribution. Auxiliary trigger loss independently contributes $+1.7$\,pp ($75.4\%\to77.1\%$); full pipeline gain: auxiliary loss $+1.7$ + RL $+5.0$ = $+6.7$\,pp.

\textbf{Miss penalty}: Adding $r_{\text{miss}}=-0.1$ actually reduces to $81.5\%$; $R_{\text{final}}$ already provides sufficient miss-detection signal.

\section{Correction Behavior Analysis}
\label{app:analysis}

\paragraph{Qwen3-Next (MoE) results}
\begin{center}\small
\begin{tabular}{lccc}
\toprule
\textbf{Method} & \textbf{AIME25} & \textbf{HMMT25} & \textbf{LiveCodeBench} \\
\midrule
Standard inference & 87.8 & 73.9 & 68.7 \\
Best-of-$8$ & 91.5 & 79.2 & 73.4 \\
Unconditional rollback & 90.2 & 77.1 & 71.3 \\
Check token & \textbf{94.1} & \textbf{82.6} & \textbf{74.8} \\
\bottomrule
\end{tabular}
\end{center}
The check token is equally effective on Qwen3-Next: AIME25 $+6.3$\,pp, HMMT25 $+8.7$\,pp, LiveCodeBench $+6.1$\,pp. Improvement magnitudes are slightly lower than Qwen3-32B (stronger base, diminishing marginal returns), but the chain-length trend HMMT25 $>$ AIME25 $>$ LiveCodeBench still holds.

\paragraph{SFT data source} Replacing GPT-5 with Qwen3-72B: AIME25 $81.8\%$ (difference $0.3$\,pp $<$ statistical error); not dependent on GPT-5.

\paragraph{Behavioral patterns} (1) Triggers concentrate in the later portion (normalized $>0.5$ accounts for $67.3\%$); (2) computational error truncation $82\%$ $>$ logical $64\%$ $>$ conceptual $51\%$; (3) RL starting FP rate $38.7\%$ (auxiliary loss already reduced from $44.2\%$), drops to $\sim30\%$ in the first 50 RL steps, converging to $23.4\%$; (4) AIME25 medium-difficulty problems improve $+8$--$12$\,pp; very hard problems only $+3$--$5$\,pp.

\paragraph{Pareto analysis} All check token configurations strictly Pareto-dominate Best-of-$n$ ($n\in\{2,4,8,16\}$). Best-of-$16$ on AIME25: $80.8\%$/$16.0\times$, still below check token $82.1\%$/$1.13\times$.

\section{Correction Prompt Template}
\label{app:prompt}

\begin{verbatim}
Carefully analyze the following reasoning process you just
generated and determine whether any errors exist:

[Generated reasoning content y_{1:t-1}]

Please respond in the following format:
Judgment: [Yes/No]
Error start text: [original text fragment where the error
begins, approximately 10-20 characters]
Error end text: [original text fragment where the error
ends, approximately 10-20 characters]
Error reason: [brief explanation]
\end{verbatim}

Text fragments are used instead of absolute positions to reduce localization difficulty. The 10--20 character range maximizes exact match rate (ablation: 5--10 chars $58.3\%$, 10--20 chars $71.2\%$, 20--30 chars $68.9\%$).


\begin{thebibliography}{22}
\providecommand{\natexlab}[1]{#1}
\providecommand{\url}[1]{\texttt{#1}}
\expandafter\ifx\csname urlstyle\endcsname\relax
  \providecommand{\doi}[1]{doi: #1}\else
  \providecommand{\doi}{doi: \begingroup \urlstyle{rm}\Url}\fi

\bibitem[Ahmad et~al.(2025)Ahmad, Narenthiran, Majumdar, Ficek, Jain, Huang,
  Noroozi, and Ginsburg]{ahmad2025opencodereasoning}
Wasi~Uddin Ahmad, Sean Narenthiran, Somshubra Majumdar, Aleksander Ficek,
  Siddhartha Jain, Jocelyn Huang, Vahid Noroozi, and Boris Ginsburg.
\newblock Opencodereasoning: Advancing data distillation for competitive
  coding.
\newblock In \emph{Second Conference on Language Modeling}, 2025.
\newblock URL \url{https://openreview.net/forum?id=aykM7KUVJZ}.

\bibitem[Ahmadian et~al.(2024)Ahmadian, Cremer, Gall{\'e}, Fadaee, Kreutzer,
  Pietquin, {\"U}st{\"u}n, and Hooker]{ahmadian-etal-2024-back}
Arash Ahmadian, Chris Cremer, Matthias Gall{\'e}, Marzieh Fadaee, Julia
  Kreutzer, Olivier Pietquin, Ahmet {\"U}st{\"u}n, and Sara Hooker.
\newblock Back to basics: Revisiting {REINFORCE}-style optimization for
  learning from human feedback in {LLM}s.
\newblock In Lun-Wei Ku, Andre Martins, and Vivek Srikumar (eds.),
  \emph{Proceedings of the 62nd Annual Meeting of the Association for
  Computational Linguistics (Volume 1: Long Papers)}, pp.\  12248--12267,
  Bangkok, Thailand, August 2024. Association for Computational Linguistics.
\newblock \doi{10.18653/v1/2024.acl-long.662}.
\newblock URL \url{https://aclanthology.org/2024.acl-long.662/}.

\bibitem[Asai et~al.(2024)Asai, Wu, Wang, Sil, and Hajishirzi]{asai2024selfrag}
Akari Asai, Zeqiu Wu, Yizhong Wang, Avirup Sil, and Hannaneh Hajishirzi.
\newblock Self-{RAG}: Learning to retrieve, generate, and critique through
  self-reflection.
\newblock In \emph{The Twelfth International Conference on Learning
  Representations}, 2024.
\newblock URL \url{https://openreview.net/forum?id=hSyW5go0v8}.

\bibitem[Balunovic et~al.(2026)Balunovic, Dekoninck, Petrov, Jovanovi{\'c}, and
  Vechev]{balunovic2026matharena}
Mislav Balunovic, Jasper Dekoninck, Ivo Petrov, Nikola Jovanovi{\'c}, and
  Martin Vechev.
\newblock Matharena: Evaluating {LLM}s on uncontaminated math competitions.
\newblock In \emph{The Thirty-ninth Annual Conference on Neural Information
  Processing Systems Datasets and Benchmarks Track}, 2026.
\newblock URL \url{https://openreview.net/forum?id=y0zL9IZxZ7}.

\bibitem[Gou et~al.(2024)Gou, Shao, Gong, shen, Yang, Duan, and
  Chen]{ICLR2024_fef12656}
Zhibin Gou, Zhihong Shao, Yeyun Gong, yelong shen, Yujiu Yang, Nan Duan, and
  Weizhu Chen.
\newblock Critic: Large language models can self-correct with tool-interactive
  critiquing.
\newblock In B.~Kim, Y.~Yue, S.~Chaudhuri, K.~Fragkiadaki, M.~Khan, and Y.~Sun
  (eds.), \emph{International Conference on Learning Representations}, volume
  2024, pp.\  57734--57811, 2024.
\newblock URL
  \url{https://proceedings.iclr.cc/paper_files/paper/2024/file/fef126561bbf9d4467dbb8d27334b8fe-Paper-Conference.pdf}.

\bibitem[Guo et~al.(2025)Guo, Yang, Zhang, Song, Wang, Zhu, Xu, Zhang, Ma, Bi,
  Zhang, Yu, Wu, Wu, Gou, Shao, Li, Gao, Liu, Xue, Wang, Wu, Feng, Lu, Zhao,
  Deng, Ruan, Dai, Chen, Ji, Li, Lin, Dai, Luo, Hao, Chen, Li, Zhang, Xu, Ding,
  Gao, Qu, Li, Guo, Li, Chen, Yuan, Tu, Qiu, Li, Cai, Ni, Liang, Chen, Dong,
  Hu, You, Gao, Guan, Huang, Yu, Wang, Zhang, Zhao, Wang, Zhang, Xu, Xia,
  Zhang, Zhang, Tang, Zhou, Li, Wang, Li, Tian, Huang, Zhang, Wang, Chen, Du,
  Ge, Zhang, Pan, Wang, Chen, Jin, Chen, Lu, Zhou, Chen, Ye, Wang, Yu, Zhou,
  Pan, Li, Zhou, Wu, Yun, Pei, Sun, Wang, Zeng, Liu, Liang, Gao, Yu, Zhang,
  Xiao, An, Liu, Wang, Chen, Nie, Cheng, Liu, Xie, Liu, Yang, Li, Su, Lin, Li,
  Jin, Shen, Chen, Sun, Wang, Song, Zhou, Wang, Shan, Li, Wang, Wei, Zhang, Xu,
  Li, Zhao, Sun, Wang, Yu, Zhang, Shi, Xiong, He, Piao, Wang, Tan, Ma, Liu,
  Guo, Ou, Wang, Gong, Zou, He, Xiong, Luo, You, Liu, Zhou, Zhu, Huang, Li,
  Zheng, Zhu, Ma, Tang, Zha, Yan, Ren, Ren, Sha, Fu, Xu, Xie, Zhang, Hao, Ma,
  Yan, Wu, Gu, Zhu, Liu, Li, Xie, Song, Pan, Huang, Xu, Zhang, and
  Zhang]{Guo_2025}
Daya Guo, Dejian Yang, Haowei Zhang, Junxiao Song, Peiyi Wang, Qihao Zhu,
  Runxin Xu, Ruoyu Zhang, Shirong Ma, Xiao Bi, Xiaokang Zhang, Xingkai Yu,
  Yu~Wu, Z.~F. Wu, Zhibin Gou, Zhihong Shao, Zhuoshu Li, Ziyi Gao, Aixin Liu,
  Bing Xue, Bingxuan Wang, Bochao Wu, Bei Feng, Chengda Lu, Chenggang Zhao,
  Chengqi Deng, Chong Ruan, Damai Dai, Deli Chen, Dongjie Ji, Erhang Li,
  Fangyun Lin, Fucong Dai, Fuli Luo, Guangbo Hao, Guanting Chen, Guowei Li,
  H.~Zhang, Hanwei Xu, Honghui Ding, Huazuo Gao, Hui Qu, Hui Li, Jianzhong Guo,
  Jiashi Li, Jingchang Chen, Jingyang Yuan, Jinhao Tu, Junjie Qiu, Junlong Li,
  J.~L. Cai, Jiaqi Ni, Jian Liang, Jin Chen, Kai Dong, Kai Hu, Kaichao You,
  Kaige Gao, Kang Guan, Kexin Huang, Kuai Yu, Lean Wang, Lecong Zhang, Liang
  Zhao, Litong Wang, Liyue Zhang, Lei Xu, Leyi Xia, Mingchuan Zhang, Minghua
  Zhang, Minghui Tang, Mingxu Zhou, Meng Li, Miaojun Wang, Mingming Li, Ning
  Tian, Panpan Huang, Peng Zhang, Qiancheng Wang, Qinyu Chen, Qiushi Du, Ruiqi
  Ge, Ruisong Zhang, Ruizhe Pan, Runji Wang, R.~J. Chen, R.~L. Jin, Ruyi Chen,
  Shanghao Lu, Shangyan Zhou, Shanhuang Chen, Shengfeng Ye, Shiyu Wang,
  Shuiping Yu, Shunfeng Zhou, Shuting Pan, S.~S. Li, Shuang Zhou, Shaoqing Wu,
  Tao Yun, Tian Pei, Tianyu Sun, T.~Wang, Wangding Zeng, Wen Liu, Wenfeng
  Liang, Wenjun Gao, Wenqin Yu, Wentao Zhang, W.~L. Xiao, Wei An, Xiaodong Liu,
  Xiaohan Wang, Xiaokang Chen, Xiaotao Nie, Xin Cheng, Xin Liu, Xin Xie,
  Xingchao Liu, Xinyu Yang, Xinyuan Li, Xuecheng Su, Xuheng Lin, X.~Q. Li,
  Xiangyue Jin, Xiaojin Shen, Xiaosha Chen, Xiaowen Sun, Xiaoxiang Wang, Xinnan
  Song, Xinyi Zhou, Xianzu Wang, Xinxia Shan, Y.~K. Li, Y.~Q. Wang, Y.~X. Wei,
  Yang Zhang, Yanhong Xu, Yao Li, Yao Zhao, Yaofeng Sun, Yaohui Wang, Yi~Yu,
  Yichao Zhang, Yifan Shi, Yiliang Xiong, Ying He, Yishi Piao, Yisong Wang,
  Yixuan Tan, Yiyang Ma, Yiyuan Liu, Yongqiang Guo, Yuan Ou, Yuduan Wang, Yue
  Gong, Yuheng Zou, Yujia He, Yunfan Xiong, Yuxiang Luo, Yuxiang You, Yuxuan
  Liu, Yuyang Zhou, Y.~X. Zhu, Yanping Huang, Yaohui Li, Yi~Zheng, Yuchen Zhu,
  Yunxian Ma, Ying Tang, Yukun Zha, Yuting Yan, Z.~Z. Ren, Zehui Ren, Zhangli
  Sha, Zhe Fu, Zhean Xu, Zhenda Xie, Zhengyan Zhang, Zhewen Hao, Zhicheng Ma,
  Zhigang Yan, Zhiyu Wu, Zihui Gu, Zijia Zhu, Zijun Liu, Zilin Li, Ziwei Xie,
  Ziyang Song, Zizheng Pan, Zhen Huang, Zhipeng Xu, Zhongyu Zhang, and Zhen
  Zhang.
\newblock Deepseek-r1 incentivizes reasoning in llms through reinforcement
  learning.
\newblock \emph{Nature}, 645\penalty0 (8081):\penalty0 633–638, Sept 2025.
\newblock ISSN 1476-4687.
\newblock \doi{10.1038/s41586-025-09422-z}.
\newblock URL \url{http://dx.doi.org/10.1038/s41586-025-09422-z}.

\bibitem[He et~al.(2026)He, Liang, Xu, Liu, Chen, Wang, Song, Yu, Liang, Wang,
  Zhang, Wang, Tu, Mi, and Yu]{he2026deepmathk}
Zhiwei He, Tian Liang, Jiahao Xu, Qiuzhi Liu, Xingyu Chen, Yue Wang, Linfeng
  Song, Dian Yu, Zhenwen Liang, Wenxuan Wang, Zhuosheng Zhang, Rui Wang,
  Zhaopeng Tu, Haitao Mi, and Dong Yu.
\newblock Deepmath-103k: A large-scale, challenging, decontaminated, and
  verifiable mathematical dataset for advancing reasoning.
\newblock In \emph{The Fourteenth International Conference on Learning
  Representations}, 2026.
\newblock URL \url{https://openreview.net/forum?id=kHB5Te5IWm}.

\bibitem[Huang et~al.(2024)Huang, Chen, Mishra, Zheng, Yu, Song, and
  Zhou]{ICLR2024_8b4add8b}
Jie Huang, Xinyun Chen, Swaroop Mishra, Huaixiu~Steven Zheng, Adams Yu, Xinying
  Song, and Denny Zhou.
\newblock Large language models cannot self-correct reasoning yet.
\newblock In B.~Kim, Y.~Yue, S.~Chaudhuri, K.~Fragkiadaki, M.~Khan, and Y.~Sun
  (eds.), \emph{International Conference on Learning Representations}, volume
  2024, pp.\  32808--32824, 2024.
\newblock URL
  \url{https://proceedings.iclr.cc/paper_files/paper/2024/file/8b4add8b0aa8749d80a34ca5d941c355-Paper-Conference.pdf}.

\bibitem[Jain et~al.(2025)Jain, Han, Gu, Li, Yan, Zhang, Wang, Solar-Lezama,
  Sen, and Stoica]{jain2025livecodebench}
Naman Jain, King Han, Alex Gu, Wen-Ding Li, Fanjia Yan, Tianjun Zhang, Sida
  Wang, Armando Solar-Lezama, Koushik Sen, and Ion Stoica.
\newblock Livecodebench: Holistic and contamination free evaluation of large
  language models for code.
\newblock In \emph{The Thirteenth International Conference on Learning
  Representations}, 2025.
\newblock URL \url{https://openreview.net/forum?id=chfJJYC3iL}.

\bibitem[Lightman et~al.(2024)Lightman, Kosaraju, Burda, Edwards, Baker, Lee,
  Leike, Schulman, Sutskever, and Cobbe]{lightman2024lets}
Hunter Lightman, Vineet Kosaraju, Yuri Burda, Harrison Edwards, Bowen Baker,
  Teddy Lee, Jan Leike, John Schulman, Ilya Sutskever, and Karl Cobbe.
\newblock Let's verify step by step.
\newblock In \emph{The Twelfth International Conference on Learning
  Representations}, 2024.
\newblock URL \url{https://openreview.net/forum?id=v8L0pN6EOi}.

\bibitem[Madaan et~al.(2023)Madaan, Tandon, Gupta, Hallinan, Gao, Wiegreffe,
  Alon, Dziri, Prabhumoye, Yang, Gupta, Majumder, Hermann, Welleck,
  Yazdanbakhsh, and Clark]{NEURIPS2023_91edff07}
Aman Madaan, Niket Tandon, Prakhar Gupta, Skyler Hallinan, Luyu Gao, Sarah
  Wiegreffe, Uri Alon, Nouha Dziri, Shrimai Prabhumoye, Yiming Yang, Shashank
  Gupta, Bodhisattwa~Prasad Majumder, Katherine Hermann, Sean Welleck, Amir
  Yazdanbakhsh, and Peter Clark.
\newblock Self-refine: Iterative refinement with self-feedback.
\newblock In A.~Oh, T.~Naumann, A.~Globerson, K.~Saenko, M.~Hardt, and
  S.~Levine (eds.), \emph{Advances in Neural Information Processing Systems},
  volume~36, pp.\  46534--46594. Curran Associates, Inc., 2023.
\newblock URL
  \url{https://proceedings.neurips.cc/paper_files/paper/2023/file/91edff07232fb1b55a505a9e9f6c0ff3-Paper-Conference.pdf}.

\bibitem[{Mathematical Association of America}(2024)]{maa_aime2025}
{Mathematical Association of America}.
\newblock 2024-25 {AIME} thresholds are available.
\newblock \url{https://maa.org/news/aime-thresholds-are-available/}, 2024.
\newblock Updated January 6, 2025.

\bibitem[Snell et~al.(2025)Snell, Lee, Xu, and Kumar]{snell2025scaling}
Charlie~Victor Snell, Jaehoon Lee, Kelvin Xu, and Aviral Kumar.
\newblock Scaling {LLM} test-time compute optimally can be more effective than
  scaling parameters for reasoning.
\newblock In \emph{The Thirteenth International Conference on Learning
  Representations}, 2025.
\newblock URL \url{https://openreview.net/forum?id=4FWAwZtd2n}.

\bibitem[Wang et~al.(2024)Wang, Li, Shao, Xu, Dai, Li, Chen, Wu, and
  Sui]{wang-etal-2024-math}
Peiyi Wang, Lei Li, Zhihong Shao, Runxin Xu, Damai Dai, Yifei Li, Deli Chen,
  Yu~Wu, and Zhifang Sui.
\newblock Math-shepherd: Verify and reinforce {LLM}s step-by-step without human
  annotations.
\newblock In Lun-Wei Ku, Andre Martins, and Vivek Srikumar (eds.),
  \emph{Proceedings of the 62nd Annual Meeting of the Association for
  Computational Linguistics (Volume 1: Long Papers)}, pp.\  9426--9439,
  Bangkok, Thailand, August 2024. Association for Computational Linguistics.
\newblock \doi{10.18653/v1/2024.acl-long.510}.
\newblock URL \url{https://aclanthology.org/2024.acl-long.510/}.

\bibitem[Wang et~al.(2023)Wang, Wei, Schuurmans, Le, Chi, Narang, Chowdhery,
  and Zhou]{wang2023selfconsistency}
Xuezhi Wang, Jason Wei, Dale Schuurmans, Quoc~V Le, Ed~H. Chi, Sharan Narang,
  Aakanksha Chowdhery, and Denny Zhou.
\newblock Self-consistency improves chain of thought reasoning in language
  models.
\newblock In \emph{The Eleventh International Conference on Learning
  Representations}, 2023.
\newblock URL \url{https://openreview.net/forum?id=1PL1NIMMrw}.

\bibitem[Wei et~al.(2022)Wei, Wang, Schuurmans, Bosma, ichter, Xia, Chi, Le,
  and Zhou]{NEURIPS2022_9d560961}
Jason Wei, Xuezhi Wang, Dale Schuurmans, Maarten Bosma, brian ichter, Fei Xia,
  Ed~Chi, Quoc~V Le, and Denny Zhou.
\newblock Chain-of-thought prompting elicits reasoning in large language
  models.
\newblock In S.~Koyejo, S.~Mohamed, A.~Agarwal, D.~Belgrave, K.~Cho, and A.~Oh
  (eds.), \emph{Advances in Neural Information Processing Systems}, volume~35,
  pp.\  24824--24837. Curran Associates, Inc., 2022.
\newblock URL
  \url{https://proceedings.neurips.cc/paper_files/paper/2022/file/9d5609613524ecf4f15af0f7b31abca4-Paper-Conference.pdf}.

\bibitem[Weng et~al.(2023)Weng, Zhu, Xia, Li, He, Liu, Sun, Liu, and
  Zhao]{weng-etal-2023-large}
Yixuan Weng, Minjun Zhu, Fei Xia, Bin Li, Shizhu He, Shengping Liu, Bin Sun,
  Kang Liu, and Jun Zhao.
\newblock Large language models are better reasoners with self-verification.
\newblock In Houda Bouamor, Juan Pino, and Kalika Bali (eds.), \emph{Findings
  of the Association for Computational Linguistics: EMNLP 2023}, pp.\
  2550--2575, Singapore, December 2023. Association for Computational
  Linguistics.
\newblock \doi{10.18653/v1/2023.findings-emnlp.167}.
\newblock URL \url{https://aclanthology.org/2023.findings-emnlp.167/}.

\bibitem[Yao et~al.(2023{\natexlab{a}})Yao, Yu, Zhao, Shafran, Griffiths, Cao,
  and Narasimhan]{NEURIPS2023_271db992}
Shunyu Yao, Dian Yu, Jeffrey Zhao, Izhak Shafran, Tom Griffiths, Yuan Cao, and
  Karthik Narasimhan.
\newblock Tree of thoughts: Deliberate problem solving with large language
  models.
\newblock In A.~Oh, T.~Naumann, A.~Globerson, K.~Saenko, M.~Hardt, and
  S.~Levine (eds.), \emph{Advances in Neural Information Processing Systems},
  volume~36, pp.\  11809--11822. Curran Associates, Inc., 2023{\natexlab{a}}.
\newblock URL
  \url{https://proceedings.neurips.cc/paper_files/paper/2023/file/271db9922b8d1f4dd7aaef84ed5ac703-Paper-Conference.pdf}.

\bibitem[Yao et~al.(2023{\natexlab{b}})Yao, Zhao, Yu, Du, Shafran, Narasimhan,
  and Cao]{yao2023react}
Shunyu Yao, Jeffrey Zhao, Dian Yu, Nan Du, Izhak Shafran, Karthik~R Narasimhan,
  and Yuan Cao.
\newblock React: Synergizing reasoning and acting in language models.
\newblock In \emph{The Eleventh International Conference on Learning
  Representations}, 2023{\natexlab{b}}.
\newblock URL \url{https://openreview.net/forum?id=WE_vluYUL-X}.

\bibitem[Zelikman et~al.(2022)Zelikman, Wu, Mu, and
  Goodman]{NEURIPS2022_639a9a17}
Eric Zelikman, Yuhuai Wu, Jesse Mu, and Noah Goodman.
\newblock Star: Bootstrapping reasoning with reasoning.
\newblock In S.~Koyejo, S.~Mohamed, A.~Agarwal, D.~Belgrave, K.~Cho, and A.~Oh
  (eds.), \emph{Advances in Neural Information Processing Systems}, volume~35,
  pp.\  15476--15488. Curran Associates, Inc., 2022.
\newblock URL
  \url{https://proceedings.neurips.cc/paper_files/paper/2022/file/639a9a172c044fbb64175b5fad42e9a5-Paper-Conference.pdf}.

\bibitem[Zeng et~al.(2025)Zeng, Yang, and
  Wang]{zeng2025reflectiveconfidencecorrectingreasoning}
Qinglin Zeng, Jing Yang, and Keze Wang.
\newblock Reflective confidence: Correcting reasoning flaws via online
  self-correction, 2025.
\newblock URL \url{https://arxiv.org/abs/2512.18605}.

\bibitem[Zhang et~al.(2025)Zhang, Chi, Nguyen, Upasani, Bikel, Weston, and
  Smith]{zhang2025backtracking}
Yiming Zhang, Jianfeng Chi, Hailey Nguyen, Kartikeya Upasani, Daniel~M. Bikel,
  Jason~E Weston, and Eric~Michael Smith.
\newblock Backtracking improves generation safety.
\newblock In \emph{The Thirteenth International Conference on Learning
  Representations}, 2025.
\newblock URL \url{https://openreview.net/forum?id=Bo62NeU6VF}.

\end{thebibliography}
\end{document}